\pdfoutput=1

\documentclass[11pt]{article}

\usepackage[]{ACL2023}

\usepackage{times}
\usepackage{latexsym}

\usepackage[T1]{fontenc}

\usepackage[utf8]{inputenc}

\usepackage{microtype}

\usepackage{inconsolata}

\usepackage{graphicx}

\title{An Investigation of
Warning Erroneous Chat Translations \\in Cross-lingual Communication}

\author{Yunmeng Li$^{1}$ 
Jun Suzuki$^{1,3}$ 
Makoto Morishita$^{2}$ 
Kaori Abe$^{1}\thanks{\ \ Currently affiliated with Machine Learning Solutions Inc.}$\ 
\ Kentaro Inui$^{4,1,3}$ \\
${}^{1}$Tohoku University
${}^{2}$NTT
${}^{3}$RIKEN
${}^{4}$MBZUAI
\\
\texttt{li.yunmeng.r1@dc.tohoku.ac.jp}
}

\begin{document}
\maketitle

\begin{abstract}

Machine translation models are still inappropriate for translating chats, despite the popularity of translation software and plug-in applications.
The complexity of dialogues poses significant challenges and can hinder cross-lingual communication.
Instead of pursuing a flawless translation system, a more practical approach would be to issue warning messages about potential mistranslations to reduce confusion.
However, it is still unclear how individuals perceive these warning messages and whether they benefit the crowd.
This paper tackles to investigate this question and demonstrates the warning messages' contribution to making chat translation systems effective.

\end{abstract}

\section{Introduction}

Globalization has led to the popularity of neural machine translation~\cite{bahdanau2014neural, vaswani2017attention, gehring2017convolutional}.
Applications like Google Translate\footnote{\url{https://translate.google.com/}} 
and DeepL\footnote{\url{https://www.deepl.com/translator}} have become essential tools in people's lives~\cite{medvedev2016google, patil2014use}.
Chat software such as WeChat and LINE also integrates built-in translation features to facilitate cross-lingual communication.
Plug-in translating applications like 
UD Talk\footnote{\url{https://udtalk.jp/}} 
and Hi Translate\footnote{\url{https://bit.ly/3pWhz9T}}
have become popular as well with the rise of online communication.

However, while machine translation technologies have demonstrated sound performance in translating documents
~\cite{barrault-etal-2019-findings, barrault-etal-2020-findings, nakazawa-etal-2019-overview, ma-etal-2020-simple, maruf-haffari-2018-document}, current methods are not always suitable for translating conversations~\cite{uthus2013multiparticipant}, especially colloquial dialogues such as chats~\cite{laubli-etal-2018-machine, toral-etal-2018-attaining, farajian-etal-2020-findings, liang-etal-2021-modeling}.
When a translation system generates erroneous translations, people unable to read the other language may not recognize such errors, leading to confusion.

\begin{figure*}[t]
\centering
\includegraphics[width=\textwidth]{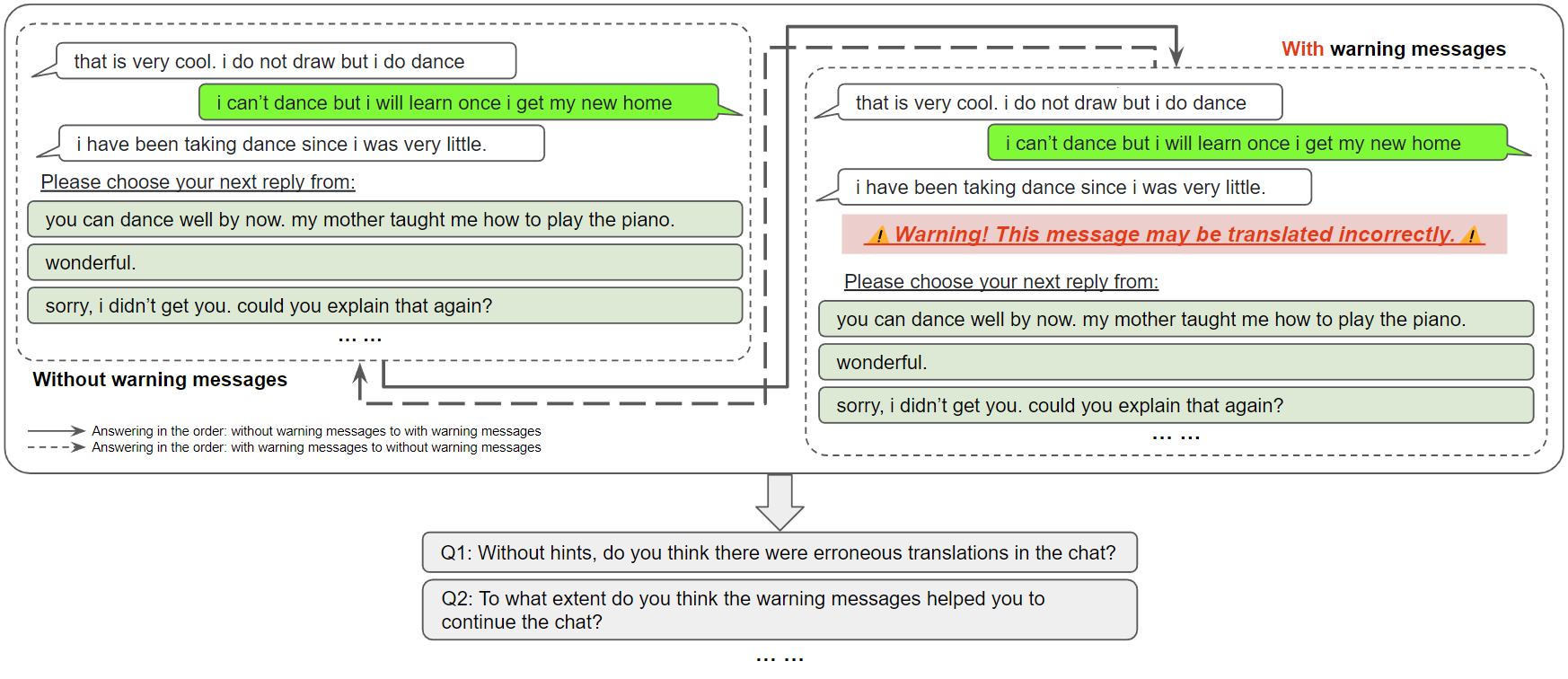}
\caption{An illustration of the designed survey. Participants will engage in two rounds of chat in the survey: one without warning messages (left) and one with warning messages (right). The content and response options are the same in both rounds. 
The order of the two rounds, either "without-with" (solid line) or "with-without" (dotted line), will be randomly assigned to participants.}
\label{fig:survey_flow}
\end{figure*}

Achieving a perfect error-free chat translation system is challenging due to the unique characteristics of chat~\cite{tiedemann-scherrer-2017-neural, maruf-etal-2018-contextual, liang-etal-2021-modeling,liang-etal-2021-towards}, making it impractical to aim for perfection.
Instead, a viable alternative approach is to enhance translation software by providing warnings about possible mistranslations to reduce confusion.
However, the perception and effects of such warning messages remain unclear.
To investigate this, we proposed to provide a warning message for erroneous translations during the cross-lingual chat and conducted a survey to explore how such warnings help people communicate.
The survey design is shown in Figure~\ref{fig:survey_flow}.
Participants engage in a simulated cross-lingual chat scenario, where they have to select the most reasonable response from three options.
Whenever a translation error occurs, a warning message is displayed.
At the end of the chat, participants answer corresponding questions regarding their perceptions of the warning messages.

We conducted the survey and collected responses through crowdsourcing. 
The results indicate that warning messages (1) are helpful in cross-lingual chats and (2) potentially encourage users to change their chat behavior.
Moreover, the survey reveals the crowd's desired features for the warning messages.
This is the first study of its kind to explore the impacts of warning users about erroneous translations in cross-lingual chat.
The findings are valuable for developing an assistant function that detects and warns users of erroneous chat translations.

\section{Related Work}

Previous studies have pointed out the potential benefits of incorporating machine translation in chat, despite its imperfections~\cite{uthus2013multiparticipant}. 
Several researchers have trained models using different methods to enhance chat translation performance~\cite{maruf-etal-2018-contextual, farajian-etal-2020-findings, liang-etal-2021-modeling}.
However, features such as ambiguity, omissions, and multi-speakers make it challenging to improve translation accuracy in chat~\cite{tiedemann-scherrer-2017-neural,liang-etal-2021-modeling,liang-etal-2021-towards}.
In contrast to existing studies of training chat translation models, we focus on acknowledging the imperfect nature of machine translation~\cite{uthus2013multiparticipant} and aim to enhance people's experience of chat translation through an alternative approach.
We propose the warning message of erroneous translation and thus improve people's experience in cross-lingual chat.
A chat translation error detector discussed in a recent study provides a binary assessment of the coherence and correctness of chat translations~\cite{li-etal-2022-chat}.
If the error detector's predictions are transformed into warning messages, our survey could be instrumental in assessing the error detector's practical effectiveness.
To the best of our knowledge, the study is the first to investigate the crowd's acceptance of such chat translation error detection tasks.

\section{Survey Design}

We propose an alternative strategy to improve translation software's performance by integrating cautionary alerts for potential mistranslations to reduce confusion.
We designed a warning message and executed a survey to evaluate its effectiveness.
Figure~\ref{fig:survey_flow} illustrates the survey process, including two simulated chat rounds: one devoid of warning messages and the other incorporating them.

\subsection{Simulated Cross-lingual Chat Scenarios}

Since dynamic real-time chats are relatively uncontrollable and high-cost, we simulated a chat scenario with a foreign partner based on chat data from Persona-chat~\cite{zhang-etal-2018-personalizing}.
In the simulation, participants are presented with three initial chat turns as historical chat logs at the beginning.
Participants choose the most contextually fitting response from the three provided options each time their scripted partners respond iteratively.
To explore the cognitive processes of individuals lacking proficiency in a foreign language, we operated under the assumption that participants would receive translated messages generated by the machine translation system from their partners.
Hence, all texts within the survey are presented to participants in their native language.

\subsection{Chat Data}

We prepared the simulated scenarios with the Persona-chat dataset, containing multi-turn chat data about various personality traits with assumed personas in English.
To ensure the quality of the data, we eliminated incoherent and unnatural chat data from Persona-chat through crowdsourcing at Amazon Mechanical Turk~\footnote{\url{https://requester.mturk.com/}}.
We defined ``incoherence'' as questions being ignored, the presence of unnatural topic changes, one speaker not addressing what the other speaker said, responses appearing to be out of order or generally difficult to follow.
We scored each chat according to the workers' answers and selected $6$ of $1,500$ chats marked as accurate and coherent by at least seven of the ten workers.
The chosen chats were used as the base of the simulated scenarios in the survey.

Similarly, we required proficient English speakers to continue the chat with given personas and topics from Persona-chat for other branching options and extended chats triggered by the options.

\subsection{Erroneous Translations}

To provide the chat data that were supposed to be erroneous translations, we translated the prepared chat data with a low-quality machine translation model that achieved a considerably low BLEU score~\cite{papineni-etal-2002-bleu} of $4.9$ on the English-Japanese chat translation evaluation dataset BPersona-chat~\cite{li-etal-2022-bpersonachat}.
Consequently, we transformed the low-quality translations twenty times through Google Translate into different languages and finally translated them back to the source language of the survey. 
To ensure the final translations could serve as erroneous translations, we manually confirmed that the texts included significant syntax issues, incorrect emotional expressions, incoherence, or other errors that led to confusion.
We designed that at least one of the three turns of the simulated chat would include erroneous translations.
We required proficient English speakers to continue the chat based on the erroneous translations to prepare the extended chat.

\subsection{Warning Messages}

We designed the warning message to notify participants of erroneous translations in the chat.
When the current text is assumed to be the erroneous translation, participants are presented with a warning message alerting them of the mistranslation, as shown in Figure~\ref{fig:survey_flow}.
We structured the warning messages into two types since receiving and sending are both essential in a conversation.
One type alerts participants of erroneous translations in the messages they received, while the other type indicates potential errors in the last message they sent.

\subsection{Corresponding Questions}

After the chat, participants are asked to answer if they notice erroneous translations without hints.
If participants answer yes, they rate their experience on two Likert Scale questions~\cite{joshi2015likert, nemoto2014likert}.
The first question assesses the extent to which the errors prevented them from continuing the chat, while the second question asks to what extent they could grasp exactly where the erroneous translations were in the message.
Participants will use 1-5 to score their perceptions, with higher numbers indicating a greater awareness or understanding of the errors.

Participants must also rate on a Likert Scale question the extent to which they think the warning helped them continue the chat. 
Further, they check the plural options of additional features they find helpful if added to the warnings.
Selectable features include:
\textit{indicating the correctness rate of the translation,
providing alternative translation suggestions,
showing specific errors in the translation,} and
\textit{suggesting the emotion of their partner.\footnote{Participants can fill in their comments or skip if they do not have any specific wanting features.}}

\section{Crowdsourcing Experiments}

We prepared the survey in English, Chinese, and Japanese to observe the possible difference between languages.
Professional translators translated the data from English to Chinese and Japanese to ensure quality.
We prepared three sets of chat data for each type of warning message and two types of warnings; hence, we provided six sets of chat and collected the responses through crowdsourcing.
We provided instructions for participants on how the chat would be presented and what they should do to attend the chat at the beginning of the task.
Participants would be acknowledged that (1) their partner would speak to them in a language other than their native language, (2) the system would translate their partners' messages and the chat would only be presented in their language, (3) they would read the chat log and choose the most reasonable of the three options, (4) the message sent to them would be displayed on the odd-numbered lines, and their answer would be displayed on the even-numbered lines.

To minimize any possible influence of showing warnings first or later, we provided each chat in two orders.
Participants answer either without warning messages first or with warning messages first. 
At the round of warning messages, we would explain the role of warning messages to participants and inform them that they could refer to the warnings to help them make choices.

We invited at least 50 participants for each order and ensured they could not join both orders through the crowdsourcing platforms' features.
Crowdworkers were unaware of the fact that there were two orders, and they did not know which order they would join.
Ultimately, we invited at least 100 participants for each set of chats.

The surveys were conducted on Amazon Mechanical Turk\footnote{\url{https://requester.mturk.com/}}
for English participants, WenJuanXing\footnote{\url{https://www.wjx.cn/}} for Chinese participants, and CrowdWorks\footnote{\url{https://crowdworks.jp/}} for Japanese participants.
Workers participated anonymously and were informed that the results would be used for academic purposes.
Classification rounds were held in advance for efficiency.

\begin{figure}[t]
\centering
\includegraphics[width=\columnwidth]{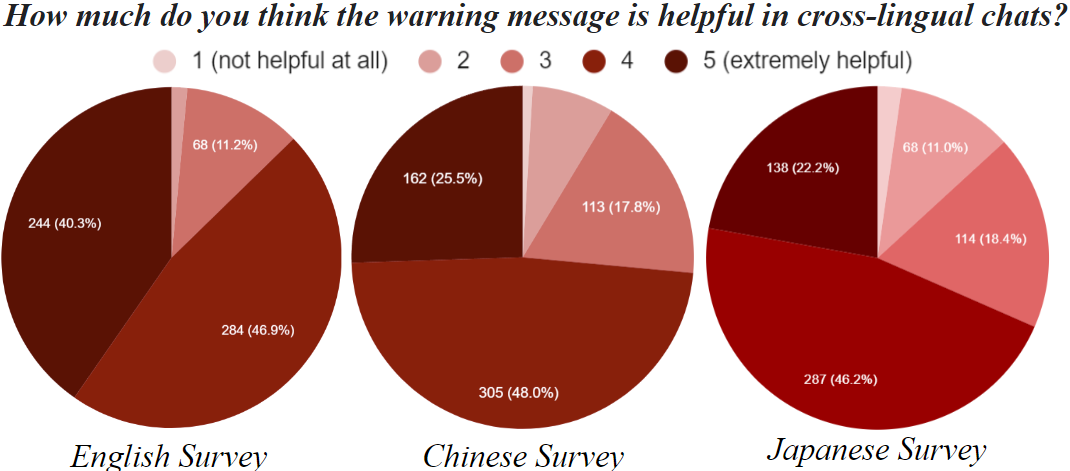}
\caption{The responses to how participants think the warning messages helped them continue the chat.}
\label{fig:total_summary}
\end{figure}

\section{Results and Analysis}

Under the different policies of crowdsourcing platforms, we finally gathered 604 English, 635 Chinese, and 621 Japanese responses.
Figure~\ref{fig:total_summary} displays the overall summaries.
Around 70\% of participants across three languages rated the warning messages as  \textit{``4 - helpful''} or above in the chat.
Most participants view the warning messages as helpful in cross-lingual chats, aligned with Likert Scale analysis~\cite{amidei-etal-2019-use}.

\paragraph{With or without warning messages}

\begin{table*}[t]
\centering
\resizebox{\textwidth}{!}{%
\begin{tabular}{lllllllllll}
\hline
 & \multicolumn{5}{l}{\textbf{Without Warning Messages First}} & \multicolumn{5}{l}{\textbf{With Warning Messages First}} \\ \hline
\textbf{English} & \multicolumn{5}{l}{\textbf{Noticing mistranslations without hints}} & \multicolumn{5}{l}{\textbf{Noticing mistranslations without hints}} \\
 & \multicolumn{5}{l}{234 of 303 (77.2\%)} & \multicolumn{5}{l}{234 of 302 (77.4\%)} \\
 & \multicolumn{5}{l}{\textbf{Considering mistranslations to be barriers}} & \multicolumn{5}{l}{\textbf{Considering mistranslations to be barriers}} \\
 & \textit{Score=1} & \textit{Score=2} & \textit{Score=3} & \textit{Score=4} & \textit{Score=5} & \textit{Score=1} & \textit{Score=2} & \textit{Score=3} & \textit{Score=4} & \textit{Score=5} \\
 & 2 & 11 & 56 & 126 & 39 & 5 & 17 & 55 & 108 & 49 \\ \hline
\textbf{Chinese} & \multicolumn{5}{l}{\textbf{Noticing the erroneous translations without hints}} & \multicolumn{5}{l}{\textbf{Noticing the erroneous translations without hints}} \\
 & \multicolumn{5}{l}{228 of 325 (70.2\%)} & \multicolumn{5}{l}{241 of 310 (77.7\%)} \\
 & \multicolumn{5}{l}{\textbf{Considering mistranslations to be barriers}} & \multicolumn{5}{l}{\textbf{Considering mistranslations to be barriers}} \\
 & \textit{Score=1} & \textit{Score=2} & \textit{Score=3} & \textit{Score=4} & \textit{Score=5} & \textit{Score=1} & \textit{Score=2} & \textit{Score=3} & \textit{Score=4} & \textit{Score=5} \\
 & 2 & 26 & 45 & 112 & 53 & 2 & 26 & 62 & 115 & 36 \\ \hline
\textbf{Japanese} & \multicolumn{5}{l}{\textbf{Noticing the erroneous translations without hints}} & \multicolumn{5}{l}{\textbf{Noticing the erroneous translations without hints}} \\
 & \multicolumn{5}{l}{175 of 321 (54.5\%)} & \multicolumn{5}{l}{158 of 300 (52.7\%)} \\
 & \multicolumn{5}{l}{\textbf{Considering mistranslations to be barriers}} & \multicolumn{5}{l}{\textbf{Considering mistranslations to be barriers}} \\
 & \textit{Score=1} & \textit{Score=2} & \textit{Score=3} & \textit{Score=4} & \textit{Score=5} & \textit{Score=1} & \textit{Score=2} & \textit{Score=3} & \textit{Score=4} & \textit{Score=5} \\
 & 3 & 21 & 29 & 89 & 33 & 1 & 17 & 29 & 86 & 25 \\ \hline
\end{tabular}%
}
\caption{The results of the questions about noticing erroneous translations without hints in the two different answering orders. Participants who answered yes to the question continued to rate the extent they considered the erroneous translations to be barriers in the chat. The higher the score was, the more confused the participant felt.}
\label{tab:with_or_without}
\end{table*}

The results of \textit{``Without hints, do you think there were erroneous translations in the chat''} based on the order in which participants answered the survey are listed in Table~\ref{tab:with_or_without}.
The percentages of noticing erroneous translations without hints remain consistent, regardless of participants answering with warning messages first or after.
Hence, we conclude that the impact of answering orders on the crowds appears minimal.
Moreover, considering a score greater or equal to 4 suggests the positivity of a Likert Scale question, we conclude that most participants who noticed erroneous translations also considered those errors as obstacles.

It is worth noting that while the English and Chinese results are relatively similar, Japanese results differ slightly.
The recognition of erroneous translations without hints is notably lower in Japanese than in English and Chinese contexts.
Participants' feedback suggests this may be related to Japanese linguistic specificity in ``omission.''
Participants considered erroneous translations as omissions, aligning with Japanese conversational patterns where subjects or objects are often omitted.
The warning messages helped them realize that the expression was not omitted but errors for the better continuation of the chat.

Additionally, English and Chinese participants also remarked that the warnings clarified unusual expressions as translation errors rather than humor or slang.
The feedback helped state the usefulness of warning messages and the consideration for future differentiation between translation errors and humorous terms or buzzwords.

\paragraph{Impact of warning messages on modifying user's chat behavior}

\begin{figure}[t]
\centering
\includegraphics[width=\columnwidth]{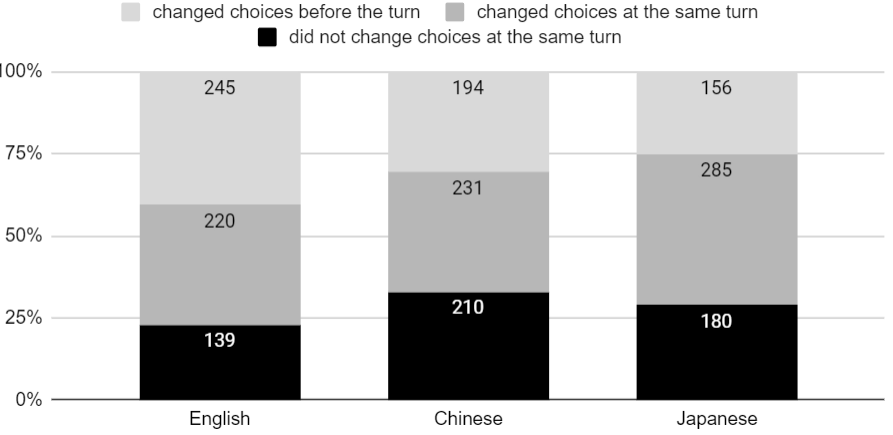}
\caption{The results that whether participants changed their choices with the help of warning messages.}
\label{fig:warning_influence}
\end{figure}

We analyzed participants' choices in relation to warning messages, categorizing them into three cases: 
(1) entered the same scenario in both the round with warnings and the round without warnings and did not change their choices,
(2) entered the same scenario in both rounds and changed their choices, and 
(3) did not change their choices due to entering other branches in advance.
We believe that the first case demonstrates that participants were not influenced by warnings, while the second case shows that they were influenced.
In the third case, although it is impossible to compare whether participants changed their choices in the same scenario since they changed earlier, we still view it as an indirect influence due to the equivalence between having no warning messages and having no erroneous translations.
Indeed, 103 participants stated they changed their choices as they ensured there were no erroneous translations.

Survey results shown in Figure~\ref{fig:warning_influence} indicate that approximately 25\% participants remained unchanged, while about 75\% changed their choices, either directly or indirectly, due to the warning messages.
We confirm that the participants were genuinely influenced by warning messages and participated in the subsequent feedback.

\paragraph{Warnings on the received messages or the sent messages}

\begin{figure}[t]
\centering
\includegraphics[width=\columnwidth]{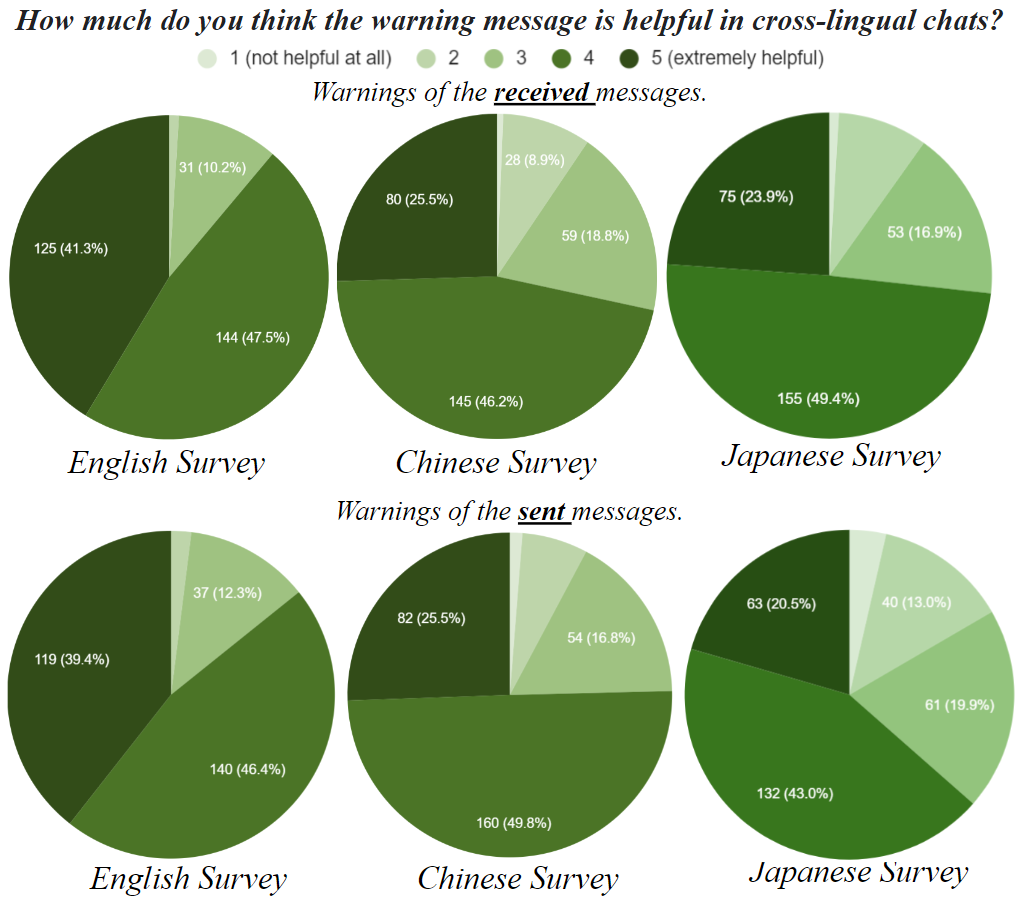}
\caption{The responses to how participants think the warnings of the \textbf{received/sent} messages helped them continue the chat.}
\label{fig:received_sent}
\end{figure}

The collected responses of different types of warning messages are summarized in Figure~\ref{fig:received_sent}.
Regardless of whether the warning messages indicated translation errors in the message received or sent, over 60\% of the participants found the warning messages helpful (rating with a score-4 or higher) in all three languages.

\paragraph{Expected features of the warning message}

\begin{figure}[t]
\centering
\includegraphics[width=\columnwidth]{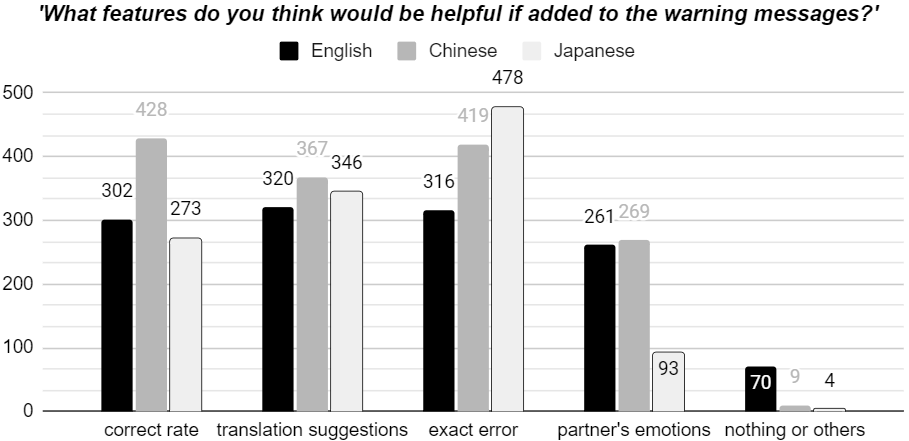}
\caption{The results about expected additional features to the warning messages.}
\label{fig:features}
\end{figure}

The results of expected additional information of the warning message are presented in Figure~\ref{fig:features}.

Chinese and Japanese participants showed a greater expectation for warning messages to indicate the exact error of their partners' messages.
In addition, Chinese participants prefer to know the correct rate.
Feedback from participants indicated that the correctness rate would better assist them in determining whether they needed to reinterpret. 
Japanese participants consider having other translation suggestions as references.
English survey participants voted on all the listed features on average, but knowing their partner's emotions were still lower than others.
In summary, to enhance the warning messages, the focus may better be on highlighting the exact errors in the translations.

\section{Conclusions}

We conducted a survey to investigate the effectiveness of warning about possible mistranslations in chat as an alternative approach to enhance the experience of cross-lingual communication.
Through crowdsourcing, we collected responses and concluded that such warning messages are helpful.
By comparing the participants' choices with and without warning messages, we found that the warning messages did encourage participants to change their behaviors.
We also found the crowd expects the warning message to (1) show the specific error in the translation, (2) indicate the correctness rate of the translation, and (3) provide alternative translation suggestions.

This survey is the first to explore the effects of warning about erroneous translations in cross-lingual chat, providing valuable insights for developing an assistant function that detects and warns people of erroneous chat translations.

\section*{Limitations}

During the survey design phase, diligent measures were taken to minimize potential leading effects on the participants' judgment by randomly switching the order and neutralizing the questioning style.
Despite the conscientious efforts, we must acknowledge the inherent challenges in completely eliminating all influences on the people who participated in the survey.
With this realization, we recognize the need for further optimization to guarantee the fairness and validity of the responses.
Refinement is warranted to minimize the biases further.

\section*{Ethics}
The crowdsourcing survey employed in this study adheres to stringent ethical guidelines to ensure participant privacy and data protection.
The survey design deliberately avoids collecting any personally identifiable information from the participants.
No restrictions or enforcement of work hours were imposed upon participants, thereby eliminating undue influence or coercion.
Given the absence of personal data collection and voluntary participation, the data is not subject to ethics review at the organization.
Consequently, the survey design and data collection procedures adhere to the ethical standards and regulations governing research practices.

\section*{Acknowledgements}
This work was supported by JST (the establishment of university fellowships towards the creation of science technology innovation) Grant Number JPMJFS2102, JST CREST Grant Number JPMJCR20D2 and JST Moonshot R\&D Grant Number JPMJMS2011 (fundamental research).
The crowdsourcing was supported by Amazon Mechanical Turk (\url{https://www.mturk.com/}), WenJuanXing (\url{https://www.wjx.cn/}) and Crowdworks (\url{https://crowdworks.jp/}).

\bibliography{anthology,custom}
\bibliographystyle{acl_natbib}


\end{document}